# Modularity in Biological Evolution and Evolutionary Computation

A.V. Eremeev[1,2], A.V. Spirov[2,3]

[1]Dostoevsky Omsk State University, Omsk, Russia

[2]The Institute of Scientific Information for Social Sciences RAS, Moscow, Russia

[3]I.M. Sechenov Institute of Evolutionary Physiology and Biochemistry RAS, St. Petersburg, Russia

**Abstract.** One of the main properties of biological systems is modularity, which manifests itself at all levels of their organization, starting with the level of molecular genetics, ending with the level of whole organisms and their communities. In a simplified form, these basic principles were transferred from the genetics of populations to the field of evolutionary computations, in order to solve applied optimization problems. Over almost half a century of development in this field of computer science, considerable practical experience has been gained and interesting theoretical results have been obtained. In this survey, the phenomena and patterns associated with modularity in genetics and evolutionary computations are compared. An analysis of similarities and differences in the results obtained in these areas is carried out from the modularity view point. The possibilities for knowledge transfer between the areas are discussed.

**Аннотация.** Одним из основных свойств биологических систем является модульность, проявляющаяся на всех уровнях их организации, начиная с молекулярно-генетического, заканчивая целыми организмами и их сообществами. В упрощенной форме эти базовые принципы были перенесены из генетики популяций в область эволюционных вычислений с целью решения прикладных задач оптимизации. За почти полувековое развитие этой области компьютерных наук накоплен значительный практический опыт и получены интересные теоретические результаты. В настоящем обзоре сопоставляются явления и закономерности, связанные с модульностью в генетике и эволюционных вычислениях. С точки зрения модульности проводится анализ сходства и различия результатов, полученных в этих областях исследований, обсуждаются возможности обмена знаниями между ними.

**Ключевые слова:** эволюционный модуль, кроссинговер, домен белка, смешиваемость, генетический алгоритм, трансфер знаний

**Введение**

Одним из факторов, повышающих вероятность благоприятных рекомбинаций и повышающих устойчивость приспособленности к локальным изменениям в генотипе, является модульность (Ратнер, 1992; Livnat et al, 2008; Schlosser, Wagner, 2004). В биологической литературе модули понимаются как подсистемы, характеризующиеся высокой степенью интеграции во внутренних связях и значительной автономностью в связях внешних (Schlosser G., Wagner, 2004). Выделяются три аспекта модульности: *модульность развития, морфологическая модульность и эволюционная модульность* (Callebaut, 2005). Несколько неформально модуль развития может определяться как подсистема, проявляющая некоторое относительно автономное поведение (Von Dassow, Munro, 1999). Как отмечено в (Muller, Wagner, 1996), по мере расширения познаний о молекулярных механизмах развития у самых разных организмов, становятся известны все новые сохраняющиеся механизмы. Некоторые из этих примеров показывают, что консервативные молекулярные механизмы могут использоваться в радикально разных контекстах развития, что позволяет предположить, что механизм развития состоит из модульных единиц, рекомбинирующихся между собой в процессе эволюции.

На морфологическом уровне модульность характеризует структуру и функцию конкретных частей или элементов организмов, таких как передняя конечность млекопитающих или модульные структуры скелетов животных. Однако, поскольку морфологические модели организации возникают в онтогенезе, морфологическая модульность может рассматриваться также как аспект модульности развития (Eble, 2005). Эволюционный модуль может быть определен на языке отображений генотип-фенотип (genotype-phenotype mapping), как набор фенотипических признаков, высоко интегрированных фенотипическими эффектами генов, определяющих их, и относительно изолированных от других подобных множеств признаков за счет незначительности плейотропных эффектов (Wagner, Altenberg, 1996). Именно эволюционный аспект модульности находится в центре внимания в настоящей статье. С этой точки зрения проводится анализ сходства и различия результатов, полученных в генетике и эволюционных вычислениях с целью выявления возможностей обмена знаниями между этими областями, в каждой из которых накоплен значительный объем знаний и подходов к пониманию эволюционной модульности. Более детальные обзоры по модульности *в биологии* могут быть найдены, например, в (Bornberg-Bauer, Albà, 2013; Callebaut, 2005; Lorenz et al, 2011; Schlosser G., Wagner, 2004).

**Эволюционные алгоритмы**

Эволюционные алгоритмы (ЭА) к числу которых относятся генетические алгоритмы (ГА), эволюционные стратегии, алгоритмы генетического программирования (ГП) и др., берут начало в работах (Ивахненко, 1971, Фогель, Оуэнс, Уолш, 1969; Holland, 1975), где было предложено моделировать процесс биологической эволюции с целью решения задач оптимизации и адаптации, а также создания систем искусственного интеллекта. Характерной особенностью ЭА является имитация процесса эволюционной адаптации биологической популяции к условиям окружающей среды, при этом особи соответствуют пробным точкам в пространстве решений задачи оптимизации (называемым *фенотипами*), а приспособленность особей определяется значениями целевой функции задачи оптимизации[1]. Пробные решения в популяции ЭА кодируются как последовательности символов некоторого алфавита, называемые *генотипами* (в ГП генотипами являются теоретико-графовые деревья, вершины которых помечены символами).

Принципы наследственности, изменчивости и отбора в эволюционных алгоритмах реализуются при построении новых решений-потомков посредством рандомизированных процедур (операторов), модифицирующих полученные ранее пробные решения подобно процессам мутации и кроссинговера (рекомбинации) в живой природе. Особям, имеющим преимущество по приспособленности, даются бо'льшие шансы быть выбранными в качестве родительских решений. Например, *($\mu,\lambda$)-селекция,* один из наиболее простых операторов селекции в ЭА, аналогичен массовому отбору в растениеводстве и животноводстве: из популяции численностью $\lambda$ отбираются $\mu$ наиболее приспособленных особей и родительские генотипы равновероятно выбираются среди них. При действии другого широко известного оператора *пропорциональной* селекции вероятность отбора особи пропорциональна ее приспособленности. На каждой итерации генетического алгоритма (Goldberg, 1989; Holland, 1975; Vose, 1999) с помощью рандомизированных операторов пропорциональной селекции, мутации и рекомбинации строится новая популяция. Операторы мутации и рекомбинации в упрощенном виде моделируют процессы мутации с заменой нуклеотидов и одиночного мейотического кроссинговера. Численность популяции фиксирована от начала работы алгоритма до конца. Рис. 1. иллюстрирует построение пары потомков в процессе генерации популяции $P(t+1)$ на основе текущей популяции $P(t)$. Обозначения операторов: S – селекция, R – кроссинговер, M – мутация. Начальная популяция $P(1)$ строится случайным образом.

---

[1] Под целевой функцией в математической оптимизации понимается функция с вещественными значениями, определенная на множестве решений задачи оптимизации. Последняя заключается в отыскании решения, на котором достигается максимум или минимум целевой функции.

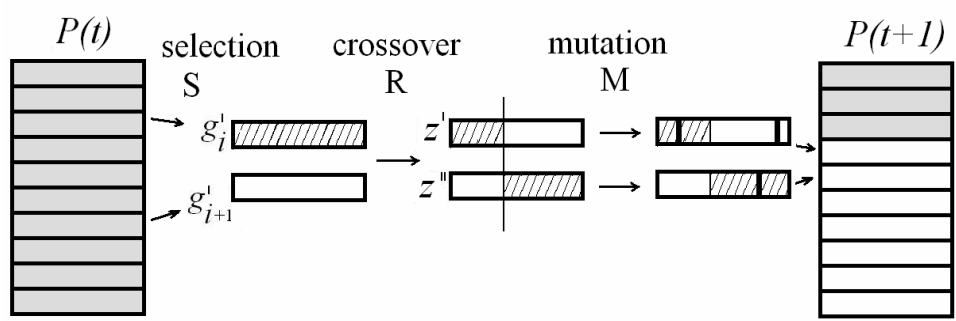

Рисунок 1. Построение пары потомков в ГА в процессе построения очередной популяции.

Подробное описание классического генетического алгоритма (Goldberg, 1989) приводится в приложении.

Ввиду простоты адаптации вычислительных схем ЭА, эти методы активно применяются для решения задач оптимизации, возникающих в управлении, планировании, проектировании, распознавании образов и других областях. Эволюционные алгоритмы, и ГА в частности, имеют многочисленные варианты, различающиеся конкретной реализацией операторов селекции, кроссинговера и мутации (De Jong, 2006). В теории ЭА, как правило, рассматриваются вопросы трудоемкости отыскания наилучшего возможного генотипа в процессе работы некоторого ЭА на выбранном классе задач. При этом в первую очередь интересуются средним числом пробных решений до первого получения оптимального генотипа, в зависимости от размерности задачи (Neumann, Witt 2010). Исследуемые классы задач могут содержать примеры сколь угодно большой размерности, как это принято в теории вычислительной сложности

(Гэри, Джонсон, 1982). Для анализа времени первого достижения оптимального генотипа (runtime analysis), как правило, используются такие методы теории вероятностей, как цепи Маркова, мартингалы, случайные процессы со сносом, стохастическое доминирование и др.

Постановка вопросов и методы исследования в теории ЭА и в популяционной генетике существенно различаются. В частности, в отношении биологической эволюции, как правило, не так важно достижение оптимума приспособленности, важнее получение достаточно приспособленных генотипов, устойчивых к возможным изменениям окружающей среды. Тем не менее, процессы, исследуемые в обеих областях, имеют много общего. В частности, эти процессы подчиняются принципам наследственности, изменчивости и отбора, соответствующим дарвиновской теории (селекционизм). На основе общности исследуемых процессов авторами (Paixão et al, 2015) предложена общая схема формального описания широкого класса эволюционных процессов, позволяющая сопоставлять аналогичные между собой модели в популяционной генетике и эволюционных вычислениях и дающая основание для переноса результатов между этими областями.

**Генетические алгоритмы в вычислительной биологии**

Несмотря на то, что разработки ЭА были вдохновлены идеями дарвиновской эволюции (в самых их общих чертах, так что можно говорить об идеях обобщенного селекционизма; сравни (Dawkins, 1983)), сходства и различия компьютерной и биологической эволюций требуют обсуждения. По совокупности концепций и наработок в область исследований ЭА резоннее всего сопоставлять с теорией и практикой эволюции биологических макромолекул (ДНК, РНК и белка). Особенно иллюстративны примеры направленной (искусственной) эволюции небольших по длине молекул ДНК, РНК и полипептидов в экспериментальных подходах типа SELEX (Gopinath, 2007; Chai et al., 2011). Здесь, как и в ГА, в начале эксперимента создается случайная популяция «особей» (обобщение понятия особи существенно для обобщенного селекционизма). В рассматриваемых экспериментальных подходах это множества биомолекул заданной длины, различающиеся своими последовательностями мономеров. С формальной точки зрения такие молекулы описываемы как последовательности букв из 4х буквенного (ДНК, РНК) или 20ти-буквенного (полипептиды) алфавита, тогда как в теории ЭА чаще всего рассматривают бинарные последовательности, но это не является критичным ограничением.

Все молекулы-особи начальной популяции подвергаются процедуре (количественной) оценки их на близость по характеристикам к искомой молекуле. Например, имеет ли она способность узнавать и связываться с заданной молекулой-мишенью (и какова специфичность и сила такого связывания). Молекулы, демонстрирующие наибольшую степень аффинности к мишени отбираются для продолжения эксперимента. Их далее умножают в числе (стадия «размножения»), используя такие экспериментальные процедуры размножения, которые делают копии отобранных молекул с определенным процентом ошибок в воспроизведении их последовательностей. Эта стадия соответствует мутации в ГА. Рекомбинация молекул тоже возможна в экспериментах по направленной эволюции, хотя она технически много сложнее, чем точечные мутации (Lutz, and Benkovk, 2008; Stebel et al., 2008). Далее, обновленная популяция вновь подвергается отбору молекул на аффинность к мишени. Отобранные лучшие молекулы вновь умножаются и мутируют. Такой цикл выполняется до тех пор, пока не будет найдена молекула требуемых характеристик.

В последние два десятилетия активно развивалась такая системно-биологическая область, как моделирование эволюции генов и генных сетей (Колчанов и др., 2008, гл. 4,5; Segal et al, 2003). Близка к ней по идеологии и подходам такая область как эволюционный дизайн генов и генных сетей (Jostins, Jaeger, 2010; Spirov, Holloway, 2013). Эти области зачастую относят к приложениям ГА к современным проблемам биологии. Степень упрощения и формализации организации генов и их функций здесь чрезвычайно высоки. Однако выводы из работ по эволюции генных сетей in silico и по эволюционному дизайну генов делаются с биологических и с эволюционно-биологических позиций.

В работах, где в рамках имитационных моделей изучается эволюция ГРС с целью изучения их эволюции в биологии (а не с целью поиска оптимальной модели, например методом фиттинга к наблюдаемым данным) (Spirov, Holloway, 2013; 2016; Payne, Moore, Wagner, 2014), процедуры рекомбинации, мутации и селекции, как правило, заимствуются из ГА. Это делается из тех соображений, что перечисленные процедуры в ГА представляют собой достаточно простые абстракции соответствующих реальных биологических процессов. Именно этим обосновывается использование ГА, а не других эвристических алгоритмов, известных в области дискретной оптимизации. ГА для моделирования биологической эволюции ГРС имеют свою специфику. Разными авторами для конкретных целей исследований ГА расширяются так, чтобы в достаточно абстрактной форме описать отображение генотипа в фенотип (см., например, (Ciliberti et al., 2007)), онтогенез (см., например, (Clune et al., 2012), диплоидность хромосомного набора (см., например, Shabash, Wiese, 2012), феномен эпистаза (см., например, (Sanjuan, Nebot, 2008; Draghi, Plotkin, 2013)) и т.д. Некоторые из этих расширений используются и в ГА для решения задач оптимизации, что происходит зачастую параллельно и независимо от реализации расширений в эволюционно-биологических моделях (Spirov, Holloway, 2016).

Более общо, можно сказать, что ЭА представляют собой весьма упрощенные и весьма формализованные описания некоторых (возможно ключевых) процессов и механизмов биологической эволюции. Критично то, что степень упрощения и формализации весьма высока и перенос заключений и выводов в областях компьютерной эволюции на реальную биологическую эволюцию приходится делать с осторожностью. Однако значительный опыт количественного анализа весьма разнообразных процессов и механизмов эволюционного поиска в ЭА безусловно заслуживает пристального изучения биологами-эволюционистами, особенно в области молекулярной биологической эволюции.

## 2. Эволюционная модульность

Анализ генотипов в популяциях различных видов животных показывает неоднородность в их структурной организации: наряду с полиморфными локусами имеются и такие, в которых отсутствует как индивидуальная, так и географическая изменчивость, так называемые мономорфные локусы (Алтухов, 2003). Межвидовые различия наиболее четко прослеживаются именно по мономорфным локусам. При анализе возникновения многих таксономических групп отмечается важная эволюционная роль *сальтационных* реорганизаций генетического материала (Алтухов, 2003, Carson, 1975). Биологический смысл этих перестроек состоит в том, что они скачком переводят значительную часть генов в мономорфных локусах в гетерозиготное состояние, тем самым, создавая качественно новые возможности адаптации популяции. Однако в схеме сальтационного видообразования имеется присутствует открытая проблема, связанная с тем, что такого рода изменчивость находится под постоянным контролем отсекающего отбора, а крупные благоприятные мутации (например, полная перестройка гена, кодирующего некоторый белок) являются маловероятными.

В работе (Ратнер, 1992) в развитие идеи (Ohno, 1970) предложен блочно-модульный принцип организации и эволюции молекулярно-генетических систем управления, дающий более правдоподобные сценарии возникновения новых белков и других биомолекул. Согласно этому принципу, эволюция генов, РНК, белков, геномов и молекулярных систем управления на их основе шла путем комбинирования блоков (модулей), причем модулями, из которых составлялись вновь возникающие молекулярно-генетические системы, служили уже функционирующие макромолекулярные компоненты. При этом дублирование генов является основным источником эволюционных инноваций, поскольку позволяет одной копии гена мутировать и исследовать генетическое пространство, в то время как другая копия продолжает выполнять исходную функцию.

Модели процесса эволюции зачастую неявно предполагают, что одна мутация для дублированного гена может дать новое свойство, увеличивающее приспособленность генотипа. Однако некоторые белковые свойства, такие как наличие дисульфидных связей или сайтов связывания лиганда, требуют участия двух или более аминокислотных остатков, которые могут потребовать нескольких мутаций. В работе (Behe, D. Snoke, 2004) моделируется простейший вариант эволюции таких белковых функций в дублированных генах. Авторы заключают, что, хотя генная дупликация и точечная мутация и могут быть эффективным механизмом исследования некоторой окрестности в генетическом пространстве, когда даже одиночные мутации приводят к увеличению приспособленности, тем не менее этот простой путь является проблематичным для развития новых функций, когда требуются множественные мутации. Таким образом, требуется рассматривать более сложные пути, возможно, связанные со вставкой, делецией, рекомбинацией или другими механизмами. Более того, эволюция регуляторных и функциональных сетей (включая сигнальные пути) зачастую требует более одной благоприятной мутации одновременно в разных генах.

**Модульность в биомолекулах**

В биомолекулах (РНК, белок) модули обычно называют доменами. Домен белка определяют как элемент третичной структуры белка, представляющий собой достаточно стабильную и независимую подструктуру, фолдинг (укладка) которой проходит независимо от остальных частей белка (Wetlaufer, 1973). Сходные по структуре домены встречаются не только в родственных белках, но и в совершенно разных. В развитие этой концепции домены определяют как компактные (относительно автономные) единицы структуры и функций, способные укладываться автономно. Имеются основания трактовать домены как единицы эволюции (Bork, 1991; Richardson, 1981). Размеры доменов чрезвычайно варьируют, но средний домен имеет порядка 100 аминокислотных остатков. Число известных доменов составляет много тысяч и продолжает расти (сравни (Neduva, Russell, 2005; Finn et al., 2016)).

Несколько доменов могут формировать мультидоменный и (часто) мультифункциональный белок (Chothia, 1992), поэтому можно говорить о генетической мобильности доменов (Davidson et al., 1993). В мультидоменном белке домены могут выполнять свои функции автономно, а могут функционировать в комплексе с остальными доменами. Соответственно, домены мультидоменного белка могут служить структурными блоками белковых комплексов (как например актомиозины), а могут выполнять специфические каталитические функции (энзимы) или узнавать и специфически связываться с определенными последовательностями нуклеиновых кислот (транскрипционные факторы). Многие домены мультидоменных белков эукариот являются независимыми (однодоменными) белками у прокариот (Davidson et al., 1993).

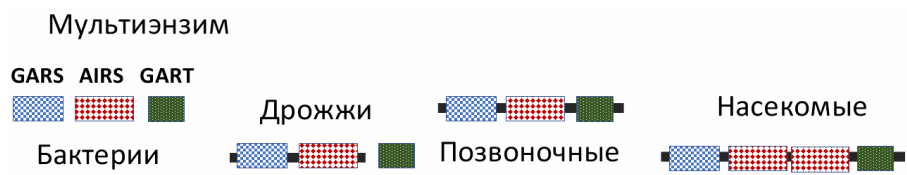

Рисунок 2. GAR-синтетазы, AIR-синтетазы и GAR трансформилазы в эволюции (Henikoff et al., 1997)

Так, например, у позвоночных имеется фермент GART (GARs-AIRs-GARt - trifunctional purine biosynthetic protein adenosine-3), состоящий из GAR-синтетазы (GARs), AIR-синтетазы (AIRs) и GAR трансформилазы (GARt) (см. рис. 2). У насекомых такой мультиэнзим включает не три, а четыре домена: домен AIRs дублирован. У дрожжей домен GARt является отдельным энзимом, тогда как пара GARs-AIRs уже объединена в один мультиэнзим. У бактерий же охарактеризованы три отдельных энзима GARs, AIRs и GARt (Henikoff et al., 1997). Эволюция в данном, достаточно типичном для мультиэнзимов случае, представима как преимущественно последовательное формирование все более сложных мультиэнзимов (ди-, три- и тетра-энзимы) из изначально относительно простых энзимов прокариот. Генетические механизмы, вовлеченные в формирование мультидоменных протеинов, включают такие масштабные реорганизации генетического материала, как инверсии, транслокации, делеции, дупликации, гомологичную рекомбинацию (Bork et al., 1992). Так как домены достаточно «автономны» в формировании своей структуры и выполнении своей функции, с помощью генной инженерии можно «пришить» к одному из белков домен, принадлежащий другому (создав таким образом белок-химеру). Такая химера при удаче будет совмещать функции обоих белков.

Так же как и для протеинов, модульность и мотивы в макромолекулах РНК (и ДНК) привлекают внимание многих исследователей. Особо подчеркивается многоуровневость модульности РНК ((Grabow et al., 2013; Grabow, Jaeger, 2013), рассмотрено нами ниже).

Известные, часто встречающиеся модули охарактеризованы в молекулярных деталях (Hendrix et al., 2005; Leontis et al., 2006; Masquida et al., 2010; Grabow et al., 2013; Grabow, Jaeger, 2013). Примеры таких модулей РНК приведены на рис.3. Существенно то, что простые модули могут входить в состав более сложных доменов РНК (Hendrix et al., 2005; Grabow et al., 2013), как на рис.3.

Тема модульности РНК привлекает особое внимание в связи с гипотезой «Мира РНК» (Gilbert, 1986, Joyce, 2002). Согласно этой гипотезе, случайно появившиеся в «первичном бульоне» небольшие молекулы РНК с разными каталитическими активностями далее кобинировались с образованием составных молекул большей сложности и большим спектром энзиматической активности. Эксперименты с селекцией РНК in vitro с использованием процедур лигации случайных последовательностей и процедур шаффлинга (shuffling), в которых образцы синтетической РНК подвергались направленной селекции для выполнения искомых экспериментаторами биохимических функций, продемонстрировали возможности получения каталитических РНК (Burke, Willis, 1998). Такие экспериментальные процедуры, по мнению исследователей, напоминают гипотетические процессы в Мире РНК. Более того, эксперименты с эволюцией на компьютере, когда простые подходы из области ЭА использовались для реализации модели направленной эволюции популяций небольших молекул РНК, подтвердили вышеприведенные эксперименты in vitro (Manrubia, Briones, 2002).

Молекулы РНК из двух желаемых модулей находились быстрее в ходе эволюционного поиска, если сначала две отдельные популяции РНК подвергались селекции на желаемые последовательности (одна популяция – первый модуль, другая -второй) и в итоге найденные мотивы сшивались в составную искомую молекулу (чем если сразу велась селекция молекул с желаемой парой мотивов).

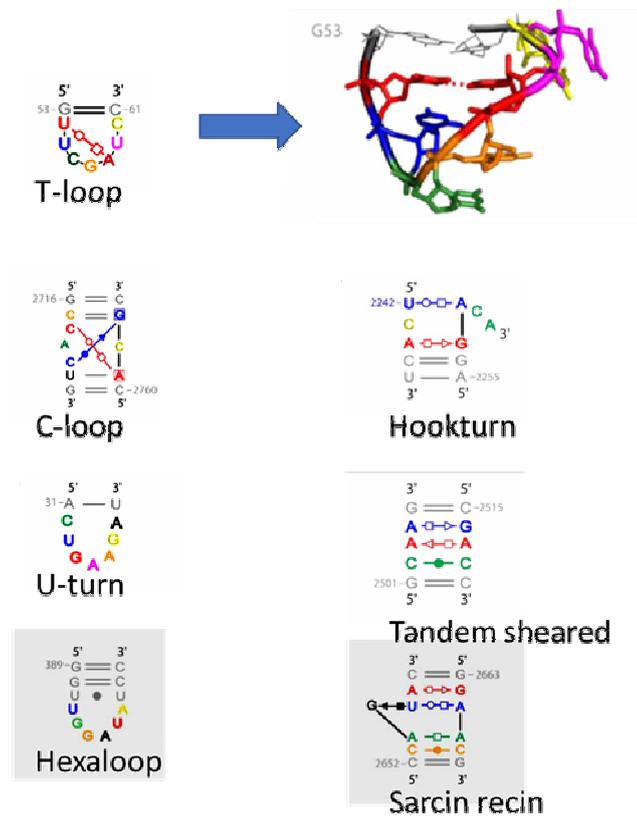

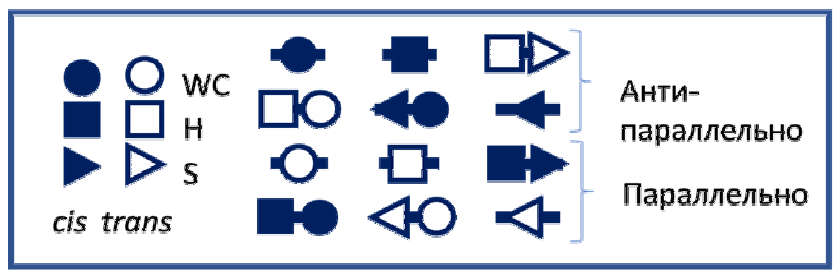

Рисунок 3. Примеры схем вторичной структуры простых мотивов РНК (Masquida et al., 2010). Для мотива T-loop приведена также третичная структура. (Обозначения схем приведены на врезке. Watson–Crick, WC, Hoogsteen, H, sugar, S. См. детали в (Masquida et al., 2010)).

В этом разделе мы вновь сталкиваемся с общей проблемой сохранения в эволюции уже найденных мотивов РНК. Здесь уместно привести наблюдения на РНК-вирусах. Как оказалось (на примере вируса иммунодефицита), области между генами РНК-вируса, содержащие структурные элементы типа шпилек, являются местами преимущественной локализации точек «перекреста» в процессах вирусного кроссинговера (Simon-Loriere et al., 2010; Simon-Loriere, Holmes, 2011). Можно предположить что наличие структурных элементов в межгенных областях РНК-вирусов обусловлено их ролью как точек перекреста, делающих более вероятным обмен интактными генами.

В недавних обзорах (Grabow et al., 2013; Grabow, Jaeger, 2013) особо подчеркивается многоуровневость (иерархичность) модульности РНК (см и рис.3). Эти биополимеры уникальны тем, что построены всего из четырех мономеров, а в основе большинства структур более высокого порядка (домены и мотивы вторичной и третичной структуры) – преимущественно комплементарные отношения между основаниями (канонические и неканонические). Эта относительная простота делает возможным формализовать основные механизмы образования и поддержания 2D и 3D доменов нуклеиновых кислот и исследовать и прогнозировать их биоинформационными и компьютерными подходами. Иерархичность модульной организации многих молекул РНК, включающих домены каждого данного уровня, составленные, в свою очередь, из субдоменов предыдущего уровня, навела ряд авторов на лингвистические аналогии (Rivas, Eddy, 2000; Jaeger et al., 2009). А именно, принципы, согласно которым составляется иерархическая структура РНК, могут быть представлены набором правил формальных грамматик, как это делается в лингвистике: части слова (субмотивы) составляются по определенным правилам в слова (мотивы), слова, в свою очередь и по своим правилам составляются предложения (мотивы или домены РНК более высокого уровня). С другой стороны, предполагается что когда мы выясним грамматику РНК, то сможем конструировать на компьютере новые молекулы РНК с желаемыми свойствами, для их последующего химического синтеза (Jaeger et al., 2009; Geary et al., 2017). (Отметим, что существенно более высокая сложность полипептидов не позволяет эффективно использовать аналогичные подходы для полипептидов.

**Модульность в генных регуляторных сетях**

Модульность генных регуляторных сетей (и более общо, модульность биологических / клеточных регуляторных сетей) – одна из широко обсуждаемых тем современной системной биологии (см например (Sole et al., 2002; Espinosa-Soto & Wagner, 2010; Clune et al., 2013; Gyorgy, Del Vecchio, 2014)). Представление ГРС графами (вершины – гены, дуги – функциональные связи между ними) наглядно иллюстрирует приводимое выше общее определение модульности, когда плотность связей внутри модуля явно выше, чем на его периферии. Каждый модуль типично включает несколько узловых генов с большим количеством связей (гены-хабы, hubs) и множество генов с небольшим числом связей. ГРС имеют тенденцию аппроксимироваться моделью масштабно-инвариантной сети (scale-free network) (Barabasi, Oltvai, 2004). Такая сеть апроксимируема графом, в котором степени вершин (число ребер, связывающих вершину с другими) распределены по степенному закону, то есть в среднем доля вершин со степенью $k$ пропорциональна $k^{-\gamma}$. Величина параметра $\gamma$ характеризует различные свойства сети, в частности, роль, которую играют гены-хабы. Одна из простых моделей, объясняющих появление масштабно-инвариантной сети известна как иерархическая модель сети, которая получается многократным применением определенных правил пошагового добавления новых слоев узлов со все меньшим числом связей к начальному кластеру генов-хабов (Barabasi, Oltvai, 2004).

Еще одна значимая особенность модульности именно в ГРС (и в клеточных регуляторных сетях) – это *мотивы* (Kashtan, Alon, 2005). Мотив характеризуют как типичный, часто встречающуюся способ функциональной связи нескольких генов. Если число генов невелико, то несложно перебрать все возможные варианты организации их в сеть. Варианты, которые встречаются много чаще чем остальные, называют мотивами. Предполагается, что мотивы реализуют собой удачные функциональные решения, пригодные для организации регуляций во многих конкретных случаях, как, например, цепь упреждения (feed-forward loop) (Kashtan, Alon, 2005). Примечательно, что удается наблюдать типичное появление определенных мотивов (например, той же цепи упреждения) в численных экспериментах по эволюции моделей ГРС (Cooper et al., 2008). Иначе говоря, как и в ряде случаев, упомянутых выше, использование подходов из области эволюционных вычислений для теоретических задач эволюционной биологии по проблемам архитектуры ГРС позволило получить согласующиеся результаты. Это пример обратного трансфера знаний из кибернетики в биологию, тогда как, в свое время эволюционные идеи биологии вдохновили кибернетиков на разработку эволюционных алгоритмов. Вместе с тем, идея «модульного дизайна» была во многом заимствована современной биологической инженерией из обычной инженерии (Hartwell et al., 1999).

**Генетическое программирование для генных сетей, операторы кроссинговера, оперирующие блоками ГРС**

Одним из широко используемых методов эволюционных вычислений является генетическое программирование. Джон Коза, автор ГП, неоднократно упоминал, что принцип дублирования с изменением элементов программ, представленных деревьями (Koza et al, 1999), был вдохновлен идеями эволюционной биологии о дупликации и последующей дивергенции дубликатов генов из книги (Ohno, 1970). Возможность представить ГРС как ориентированный граф (так что дуги, при этом имеют знак: плюс – активирующее действие, минус – репрессия) давно используется биологами: это делает наглядными как архитектуру сети, так и «мутации» сети (если под мутациями понимать добавление – удаление генов (узлов) и добавление – удаление регуляторных связей (дуг). Активно работающие в этой области лаборатории Сиггиа и Франке обозначили свое направление как эволюция *in silico* (Francois, Hakim, 2004; Francois et al., 2007; Francois, Siggia, 2010). Пример такого графического представления мутаций модели генной сети приведен на рис.4.

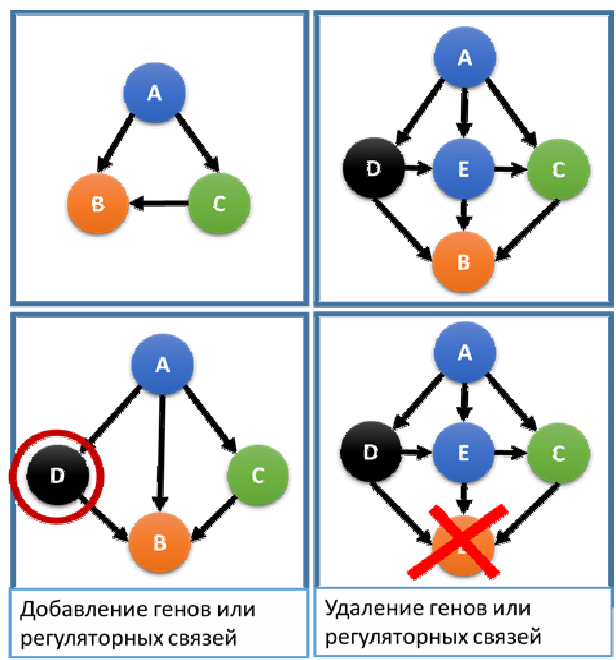

Рисунок 4. Пример представления ГРС как ориентированного графа, где вершинам соответствуют гены, а дугам соответствует регулирующее действие продукта одного гена на другой. Проиллюстрированы два основных класса мутаций: добавление (1) или удаление (2) генов или их регуляторных связей).

«Мутации» при таком представлении ГРС – это добавление или удаление генов (узлов) и добавление или удаление регуляторных связей (дуг), рис.4. Кроссинговер при таком представлении ГРС может быть реализован как обмен частей генных сетей, как это обычно реализуется в техниках ГП. Такое представление ГРС, в свою очередь, делает возможным непосредственное использование техники ГП для исследований ГРС. Среди довольно широкого круга фундаментальных и прикладных задач в области ГРС нас интересует использование техник ГП для обратной инженерии архитектуры ГП и для исследований эволюции ГРС на компьютере.

В области обратной инженерии организации ГРС, когда исходят из экспериментальных данных по генной экспрессии для вывода архитектуры сети, приложения техник ГП развиваются несколькими группами. Одна из примечательных работ в этой области была опубликована основателем ГП Дж. Козой с соавторами (Koza et al., 2000) и описывала использование ГП для обратной инженерии регуляторной сети лактозного оперона бактерии. В области приложения ГП к анализу биологических и модельных, искусственных регуляторных сетей активно работает группа Банзхафа (Banzhaf, 2003; Leier et al., 2006; Hu et al., 2014). Приложение техник ГП для исследования ГРС и их эволюции – это еще один пример обратного трансфера знаний из кибернетики в биологию. Заслуживает упоминания здесь то, что мы не нашли в области приложения техник ГП к проблемам ГРС примеров использования классических схем кроссинговера, используемых в ГП. Соответственно мы можем ожидать дальнейшего прогресса именно в области приложений таких операторов кроссинговера к проблемам ГРС. Заметим здесь также, что простота и естественность кроссинговера как переноса некоторых подграфов графа ГРС должна ускорять эффективность эволюционного поиска архитектуры ГРС по сравнению с другими эволюционными подходами и другими реализациями архитектуры ГРС. В этой связи нам представляется весьма интересным вопрос о том, возможно ли так организовать генетический материал и использовать такие генно-инженерные методы, что бы они в итоге выполняли кроссовер согласно схемам ГП.

**Модульность как фактор устойчивости**

A. Livnat, C. Papadimitriou, J. Dusho и M.W. Feldman определяют новую величину, называемую *смешиваемостью* (mixability), которая характеризует способность аллелей успешно функционировать в разных комбинациях с другими аллелями (Livnat et al, 2008) и обозначается через $\overline{M}$. По определению,

$$\overline{M} = \frac{\sum_{l \in L} \sum_{i \in l} P_{i,t} M_i}{|L|},$$

где $L$ – множество всех локусов, каждый из которых рассматривается как множество аллелей в этом локусе, $|L|$ – число всех локусов, $i \in l$ – аллель из локуса $l \in L$, $M_i$ – средняя приспособленность всех генотипов с аллелью $i$.

Используя имитационное моделирование в рамках классического популяционно-генетического подхода, авторы этой работы показывают, что в широком диапазоне условий при половом размножении селекция способствует преобладанию аллелей с высокой смешиваемостью. Иначе говоря, согласно (Livnat et al, 2008), селекция благоприятствует такому варианту аллели (или в общем случае варианту последовательности на некотором участке ДНК), который поддерживает высокую приспособленность при рекомбинации с различными родительскими генотипами. Эти исследования подтверждают интуитивное предположение Кроу и Кимуры (Crow, Kimura, 1965) о том, что половое размножение благоприятствует преумножению «хорошо смешиваемых» аллелей (good mixers), которые Кроу и Кимура охарактеризовали как аллели, которые вносят большой *аддитивный* вклад в приспособленность. Данное предположение согласуется с эмпирическими примерами. В частности, Кроу и Кимура отметили, что в случаях, когда у дрозофилы резистентность к лекарственным средствам развивалась в условиях полового размножения, вклад различных хромосом в резистентность имел аддитивный характер (Crow, Kimura, 1965; King, Somme, 1958). С другой стороны, резистентность к лекарственным средствам, развившаяся при бесполом размножении в Escherichia coli, была уменьшена последующей рекомбинацией, что говорит о значительной неаддитивности влияния разных аллелей в этом случае (Cavalli, Maccacaro, 1952). При половом размножении локусы в моделях (Livnat et al, 2008) также приобретают характерные черты эволюционных модулей по (Schlosser, 2004), т. е. каждый локус вносит свой вклад в приспособленность, мало зависящий от состояний других локусов. Механизм отбора аллелей по признаку смешиваемости, описанный в (Livnat et al, 2008), может также способствовать пониманию результатов (Misevic et al., 2006), которые обнаружили, что генетические элементы, кодирующие один и тот же признак, оказывались физически ближе друг к другу в двуполых организмах, чем в бесполых.

**Модульность в теории эволюционных алгоритмов**

Начиная с первых шагов теории ЭА, аспекты модульности исследовались в рамках теоретического анализа *схем* (Holland,1975) и *форм* (Radcliffe, 1991). В соответствии с общепринятым в эволюционных вычислениях подходом предположим, что все генотипы представляют собой строки из $n$ бит. Схемой $H$ с $q$ фиксированными позициями $j(1),j(2),...,j(q)$ называют множество генотипов

$$H=\{g=(g_1, g_2,…,g_n) \mid g_{j(1)} = h_1, g_{j(2)} = h_2,… g_{j(q)} = h_q\},$$

где $j(1)<j(2), j(2)<j(3),...j(q-1)<j(q)$. То есть, схеме $H$ принадлежат все те генотипы, у которых в локусах $j(1),j(2),..., j(q)$ находятся аллели $h_1,h_2,...,h_q$, соответственно. Число $q$ будем называть порядком схемы и обозначать через $q(H)$. Длиной $l=l(H)$ схемы $H$ называют расстояние между крайними фиксированными позициями, т.е. $l(H)=j(q(H))-j(1)$. Любой элемент пространства генотипов является частным случаем схемы порядка $q=l$. Множество всех двоичных строк $G=\{0,1\}^n$ является схемой без фиксированных позиций. Заметим, что одним генотипом могут обладать несколько особей популяции ГА. Введем обозначение $N(H,P(t))$ для числа генотипов из схемы $H$ в поколении $t$. Тогда среднее значение приспособленности на особях схемы $H$ в поколении $t$ есть

$$w(H,P(t)) := \frac{\sum_{i: g_{it} \in H} w(g_{it})}{N(H,P(t))}.$$

Рассмотрим оценку среднего числа представителей заданной схемы среди особей нового поколения классического генетического алгоритма (см. приложение), традиционно называемую *теоремой о схемах* (Goldberg,1989; Holland,1975).

***Теорема 1.*** *Пусть в классическом генетическом алгоритме при вероятности мутации $p$ и вероятности кроссинговера $p_c$ для некоторого $t$ выполнено условие $w(H,P(t)) \geq c\, w(G,P(t))$, тогда имеет место неравенство*

$$E[N(H,P(t+1))] \geq c\left(1 - \frac{l(H)p_c}{n-1}\right)(1-p)^{q(H)} N(H,P(t)),$$

*где $E[\cdot]$ обозначает математическое ожидание.*

Теорема о схемах показывает, что при выборе кодировки решений разработчик ГА должен стремиться к тому, чтобы перспективные свойства решений были бы представлены в генотипе в виде его коротких участков, называемых *строительными блоками* (СБ).

**3. Целенаправленное сохранение модульности**

Потребность в операторах кроссинговера, сохраняющих СБ была осознана в ЭА в ходе развития теории и практики применения ГА и ГП. Это происходило независимо от положения дел в современной биологии. В биоинженерии параллельно и независимо развиваются молекулярно-инженерные подходы для манипулирования с доменами в направленной эволюции биологических макромолекул (Voigt et al., 2002). В ЭА разработан целый ряд операторов кроссинговера, способных сохранять СБ (Skinner, Riddle, 2004; Zaritsky and Sipper, 2004; Li et al., 2006; El-Mihoub et al., 2006; Kameya, Prayoonsri, 2011; Umbarkar and Sheth, 2015). Мы можем ожидать, что логика этих алгоритмов подойдет для разработки генноинженерных подходов в области направленной эвлюции биомолекул, прежде всего. Кроссинговер, сохраняющий модули – СБ, требуется в ГА и ГП, в эволюционном дизайне моделей генных сетей, направленной эволюции макромолекул.

**Сохранение модульности в экспериментальной эволюции**

В начале 90х Виллем Стеммер разработал метод «перетасовки» ДНК (DNA shuffling) (Stemmer, 1994a,b). Метод предполагает высокую гомологию рекомбинируемых последовательностей. Это была, по-видимому, первая эффективная реализация генно-инженерных процедур, сходных с гомологичным кроссинговером. В обеих своих первых публикациях поэтому подходу (Stemmer, 1994a,b) Стеммер отмечает, что именно в области ГА была продемонстрирована высокая эффективность кроссинговера (в сравнении с точечными мутациями) для решения сложных задач эволюционным поиском. Новые экспериментальные молекулярно-биологические процедуры, рекомбинирующие цельные модули, разработаны в области sexual methods в белковой инженерии (Lutz, and Benkovk, 2008). Эти биотехнологические подходы предполагают выявление, «вырезание» и «перетасовку экзонов» (exon shuffling) «сшиванием» их в новые химерные молекулы, сходно с тем, как, полагают, это происходило в биологической эволюции (Stebel et al., 2008). Отметим в заключение, что методы негомологической рекомбинации не входят в набор стандартных ГА (хотя некоторые авторы развивают процедуры кроссовера, напоминающие методы негомологической рекомбинации в биотехнологии). Поэтому теоретический анализ эффективности таких методов был бы важен и для ЭА и для направленной эволюции. Примечательно так же то, что принцип экзонно-интронной организации гена вдохновил целую серию публикаций по новым техникам ЭА. Эти новые методы в ЭА были вдохновлены ожиданиями биологов, что обширные интронные вставки между экзонами, кодирующими домены, способствуют сохранению целостности доменов при рекомбинации (Nordin et al., 1997; Kouchakpour et al., 2009; Rohlfshagen, Bullinaria, 2010).

Таким образом, для сохранения модульности требуется специальные кроссинговерные механизмы. Такие механизмы разработаны в области направленной эволюции макромолекул. В этих подходах используют ферменты и особенности организации генов. Есть ли подобные механизмы в живой природе – открытый вопрос.

**Конфликт интересов.** Авторы заявляют об отсутствии конфликта интересов.

**Список литературы**

**Приложение 1**

В настоящем приложении приводится подробное описание классического генетического алгоритма (КГА), одного из наиболее известных вариантов ГА (Goldberg, 1989) и, как правило, используется система обозначений из (Paixão et al, 2015). Пусть требуется найти максимум неотрицательной целевой функции $f(x)$ на пространстве решений $X$. При использовании КГА решения $x \in X$ представляются двоичными строками фиксированной длины $n$. В литературе по эволюционным вычислениям решения из пространства $X$ принято называть *фенотипами*, а двоичные строки из множества $G=\{0,1\}^n$ – *генотипами*. Компоненты строки генотипа (биты) $g^1,g^2,…,g^n$ принято называть *генами*. Каждому генотипу $g \in G$ сопоставляется элемент множества $X$, т.е. определяется отображение генотип-фенотип $\varphi:G\to X$. Композиция отображений $w(g)=f(\varphi(g))$ называется *функцией приспособленности* и определяет «адаптированность» генотипа $g$ к задаче оптимизации функции $f$.

Популяцией численности $k$ называется вектор пространства $G^k$. Способ нумерации особей в популяции КГА не имеет значения. Популяция поколения $t$, $t=1,2,…$ будет обозначаться через $P(t)=(g_{1t},g_{2t},…,g_{kt})$. Численность популяции $k$ фиксирована от начала работы алгоритма до конца и для простоты предполагается четной. Итерацией КГА (см. рис. 1) является переход от текущей популяции $P(t)$ к следующей $P(t+1)$.

Приведем общую схему КГА. Используемые здесь рандомизированные операторы пропорциональной селекции $S_{Prop(f)}$, одноточечного кроссинговера $R_{1Point}$ и мутации $M_p$ будут описаны ниже.

**Классический генетический алгоритм**

1. **Для** $i$ **от** 1 **до** $k$ **выполнять** шаг 1.1:

1.1. Построить случайным образом генотип $g_{i1}$

2. $t:=1$

3. **Пока** не удовлетворяется условие остановки **выполнять** шаги 3.1-3.4

3.1. Выбрать набор родительских генотипов $(g'_1,g'_2,…,g'_k):= S_{Prop(f)}(P(t))$

3.2. **Для** $j$ **от** 1 **до** $k/2$ **выполнять** шаги 3.2.1-3.2.2:

3.2.1. $(z',z''):= R_{1Point}(g'_{2j-1},g'_{2j})$ (коссинговер)

3.2.2. $g_{2j-1,t+1}:=M_p(z')$, $g_{2j,t+1}:=M_p(z'')$ (мутация)

3.2.3. **Конец цикла**

3.3. $t:=t+1$

3.4. **Конец цикла**

4. Результат КГА – генотип с наибольшей приспособленностью $f(\varphi(g_{it}))$ среди найденных.

Поясним приведенную схему. На шаге 1 формируется начальная популяция $P(1)$, элементы которой генерируются в соответствии равновероятно на множестве генотипов $G$. Действие оператора пропорциональной селекции на пространстве популяций $S_{\text{Prop}(f)}:G^k \to G^k$ имеет то же значение, что и естественный отбор в природе. Каждый элемент набора родительских генотипов $(g'_1,g'_2,\ldots,g'_k)$ выбирается из популяции $P(t)$ независимо от других, при этом с вероятностью $\Pr(g_{it} \to g_l')$ в позицию $l=1,\ldots,k$ набора $(g'_1,g'_2,\ldots,g'_k)$ копируется любой генотип $g_{it}$, $i=1,\ldots,k$. По определению оператора пропорциональной селекции вероятность $\Pr(g_{it} \to g_l')$ пропорциональна $f(\varphi(g_{it}))$, т.е. $\Pr(g_{it} \to g_l') = f(\varphi(g_{it})) \bigg/ \left(\sum_{j=1}^{k} f(\varphi(g_{jt}))\right)$.

Результат одноточечного кроссинговера $(z',z''):= R_{1\text{Point}}(x,y)$ с фиксированной вероятностью $p_c$ формируется в виде
$$z':=(x_1,x_2,\ldots,x_m, y_{m+1},\ldots,y_n),$$
$$z'':=(y_1,y_2,\ldots,y_m, x_{m+1},\ldots,x_n),$$
где случайный номер координаты скрещивания $m$ выбран равновероятно от 1 до $m$-1. В противном случае (т.е. с вероятностью $1-p_c$) оба генотипа сохраняются без изменений, т.е. $z':=x$, $z'':=y$. Влияние оператора кроссинговера регулируется настраиваемым параметром $p_c \in [0,1]$.

Оператор мутации $M_p$ в каждой позиции генотипа с заданной вероятностью $p$ изменяет ее содержимое. В противном случае ген остается без изменений. Таким образом, мутация элементов генотипа происходит по схеме Бернулли с вероятностью успеха, равной $p$.

Выбор параметров численности популяции $k$, вероятностей мутации $p$ и кроссинговера $p_c$ позволяет регулировать работу КГА и настраивать его на конкретные задачи. Увеличение вероятности мутации до 0.5 превращает КГА в простой случайный перебор, имеющий весьма ограниченное применение.

Уменьшение $p$ до нуля приводит к малому разнообразию генотипов в популяции и может вызвать «зацикливание» КГА, когда на каждой итерации генерируются лишь ранее встречавшиеся генотипы. Стандартным оператором мутации считается $M_p$ при выборе $p=1/n$. Вариант описанного ГА, в котором оператор кроссинговера возвращает только один из двух генотипов $z', z''$, выбранный равновероятно, а шаг 3.2.1 имеет вид $z' := R_{1Point}(g'_{2j-1}, g'_{2j})$, $z'' := R_{1Point}(g'_{2j-1}, g'_{2j})$, называется *простейшим генетическим алгоритмом* и интенсивно исследуется в теории эволюционных алгоритмов (Vose, 1999).